\documentclass[letterpaper, 10 pt, conference]{ieeeconf}

\IEEEoverridecommandlockouts
\usepackage{microtype}

\usepackage{times}
\usepackage{amssymb}
\usepackage{amsmath}
\usepackage{amsthm}
\usepackage{graphics}
\usepackage{epsfig}
\usepackage{multirow}
\usepackage{nicefrac}
\usepackage{cases}          
\usepackage{subcaption}
\usepackage{mathtools}
\usepackage{xcolor}
\usepackage{algorithm}
\usepackage{algorithmic}
\usepackage{balance}
\usepackage{gensymb}
\usepackage{comment}
\usepackage{hyperref}
\usepackage[capitalise]{cleveref}
\usepackage{bm}

\hypersetup{hidelinks}

\DeclareMathAlphabet{\pazocal}{OMS}{zplm}{m}{n}

\newcommand{\mat}[0]{\begin{bmatrix}}
\newcommand{\mate}[0]{\end{bmatrix}}
\newcommand{\va}{\mathbf{a}}

\newcommand{\vM}{\mathbf{M}}
\newcommand{\vn}{\mathbf{n}}
\newcommand{\vp}{\mathbf{p}}

\newcommand{\vs}{\mathbf{s}}

\newcommand{\vO}{\mathbf{O}}

\newcommand{\vu}{\mathbf{u}}

\newcommand{\vx}{\mathbf{x}}
\newcommand{\vX}{\mathbf{X}}

\newcommand{\cA}{\mathcal{A}}

\newcommand{\cI}{\mathcal{I}}

\newcommand{\cO}{\mathcal{O}}

\newcommand{\cU}{\mathcal{U}}

\newcommand{\cW}{\mathcal{W}}
\newcommand{\cX}{\mathcal{X}}

\newcommand{\R}{\mathbb{R}}

\newcommand\norm[1]{\left\|#1\right\|}              
\newcommand\abs[1]{\left|#1\right|}                 

\usepackage{siunitx}

\usepackage{graphicx}
\usepackage{placeins}
\usepackage{xcolor}

\usepackage{cleveref}

\usepackage{comment}

\usepackage[sort, compress]{cite}

\usepackage[shortcuts]{extdash}

\crefrangeformat{figure}{Figures~#3#1#4~through~#5#2#6}
\crefname{figure}{Fig.}{Figures}

\title{\LARGE \bf
Where to Look Next: Learning Viewpoint Recommendations \\ for Informative Trajectory Planning}

\author{
Max Lodel, Bruno Brito, Álvaro Serra-Gómez, Laura Ferranti, Robert Babuška, Javier Alonso-Mora
\thanks{This work was  supported by the National Police of the Netherlands. All content represents
the opinion of the authors, which is not necessarily shared or endorsed by
their respective employers and/or sponsors. L. Ferranti received support from the Dutch Science Foundation
NWO-TTW within the Veni project HARMONIA (nr. 18165).}%
\thanks{The authors are with the Department of Cognitive Robotics (CoR), Delft University
of Technology, 2628CD Delft, The Netherlands,
{\tt \{m.lodel; bruno.debrito; a.serragomez; l.ferranti; r.babuska; j.alonsomora\}@tudelft.nl}. R. Babu\v{s}ka is also with CIIRC, Czech Technical University in Prague, Czech Republic.}%
\thanks{\textbf{Video:}
\tt \href{https://youtu.be/qxabfC9I66k}{https://youtu.be/qxabfC9I66k}}
}

\begin{document}
\bstctlcite{BSTcontrol}

\maketitle
\thispagestyle{empty}
\pagestyle{empty}

\begin{abstract}



Search missions require motion planning and navi-gation methods for information gathering that continuously replan based on new observations of the robot's surroundings. Current methods for information gathering, such as Monte Carlo Tree Search, are capable of reasoning over long horizons, but they are computationally expensive. An alternative for fast online execution is to train, offline, an information gathering policy, which indirectly reasons about the information value of new observations. However, these policies lack safety guarantees and do not account for the robot dynamics. 
To overcome these limitations we train an information-aware policy via deep reinforcement learning, that guides a receding-horizon trajectory optimization planner.
In particular, the policy continuously recommends a reference viewpoint to the local planner, such that the resulting dynamically feasible and collision-free trajectories lead to  observations that maximize the information 
gain and reduce the uncertainty about the environment. In simulation tests in previously unseen environments, our method consistently outperforms greedy next-best-view policies and achieves competitive performance compared to Monte Carlo Tree Search, in terms of information gains and coverage time, with a reduction in execution time by three orders of magnitude.

\end{abstract}

\section{Introduction}

Autonomous robots can play a fundamental role in gathering information in critical and dynamic scenarios, such as search and rescue \cite{Tian2020a,Niroui2019} or environmental monitoring \cite{Popovic2017, Vasilijevic2017}.
For example, robots can support human emergency responders to locate victims in challenging or dangerous terrain.
In such scenarios, environments are often unknown, and autonomous navigation methods must continuously replan actions that maximize the information gathered over long horizons. Moreover, these trajectories must be efficient with respect to time or energy costs. 

Long horizon, or \textit{non-myopic}, path planning methods for information gathering and map exploration \cite{Lauri2016, Best2019, Patten2018, Atanasov2014, Schlotfeldt2018, Cao2021, Charrow2015, Popovic2017, Meera2019} suffer from high computational cost and thus long planning times, particularly in complex, obstacle-rich environments.
To enable fast online execution, recent works have approached information gathering using deep reinforcement learning (DRL) \cite{Jeong2019, Chen2019, Zhu2018, Niroui2019, Viseras2019, Julian2019, Chaplot2020}.
In these methods, a policy learns in offline training to select an action that
maximizes the expected information gain of future observations.
The policy is usually modeled as a deep neural network that reasons about high-dimensional observations of the agent's surroundings (e.g., obstacles), or its current belief about the environment.
However, DRL-based information gathering methods do not explicitly consider constraints for collision avoidance and do not account for the robot's dynamics.

\begin{figure}[t]
    \centering
    \vspace{0.05in}
     \includegraphics[width=\columnwidth]{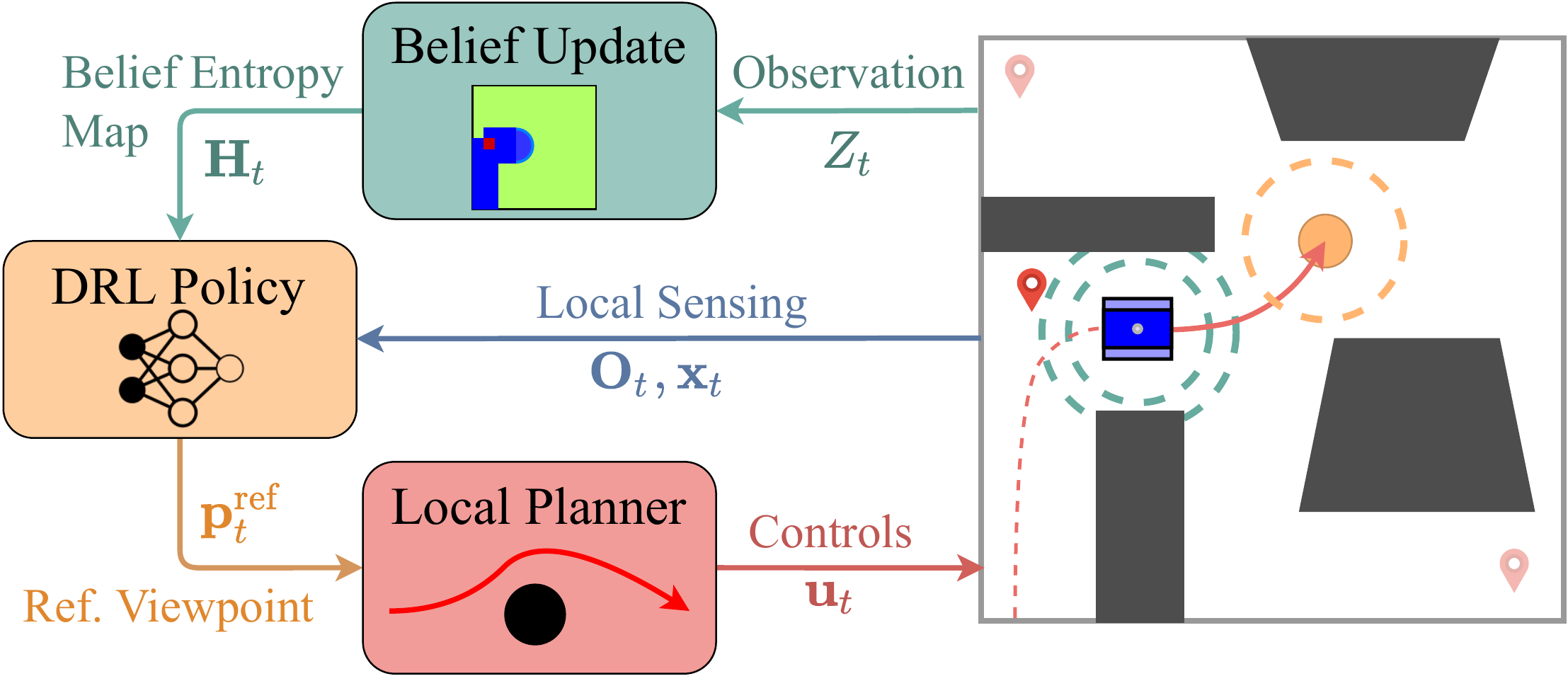}
    \caption{Conceptual overview of the proposed informative trajectory planning framework. A DRL policy recommends a viewpoint reference to a local planner, based on the robot's current belief about the environment, and local sensory information. The local planner generates a feasible trajectory and executes control commands, leading the robot (depicted in blue on the right-hand side) to reduce the uncertainty about the environment.}
    \label{fig:overview}
    \vspace{-2ex}
\end{figure}

In uncertain or dynamic scenarios, it is advantageous to employ an optimization-based local motion planner such as model predictive control (MPC), to generate dynamically feasible and collision-free trajectories and thus safe robot motion \cite{Brito2019, Zhu2019}.
%
%
The recent work of \cite{Brito2021} combined MPC with a learned subgoal policy for navigation among interacting agents.
In this paper, we propose a hierarchical framework for exploring unknown, obstacle-rich environments. Building on the idea of \cite{Brito2021}, we enhance a local motion planner with a guidance policy trained using DRL.
By training in different simulated environments, the DRL agent learns a guidance policy that maximizes information gains from future sensor observations.
In particular, the policy is trained to combine its belief about the environment with local observations of obstacles and the robot state for guiding an MPC-based motion planner by recommending a subgoal reference.
%
This \textit{viewpoint reference}, i.e., a subgoal 
leading towards informative observations,
is then used by the MPC planner to generate low-level control commands while ensuring collision avoidance and kinodynamical feasibility of the trajectory. \looseness=-1


\subsection{Related Work}
\label{sub:2_rel_work}

\subsubsection{Planning for Information Gathering}
Informative path planning (IPP) methods plan future observation poses that are expected to reduce uncertainty about the environment 
as efficiently as possible, generally at the expense of computation time.
Generally, \textit{myopic} and \textit{non-myopic} IPP methods can be distinguished. 
Myopic methods capitalize on the submodularity
property of common IPP objectives such as maximizing mutual information \cite{Singh2009}. These methods select their actions greedily either by considering the next best viewpoint at the current time step \cite{Bourgault2002, Gonzalez-Banos2002, Stachniss2005}, 
or by finding a trajectory leading to the best reachable next viewpoint \cite{Bircher2018}. 
While their computation times are generally low, these methods sacrifice efficiency in terms of time or energy required to gather information about the environment due to their short planning horizon. 





Non-myopic planning methods, in contrast, attempt to find long-horizon plans that maximize an information-related objective quantifying the cumulative information gain. 
These methods often rely on tree search algorithms \cite{Lauri2016, Patten2018, Best2019, Schlotfeldt2018, Atanasov2014}, such as Monte Carlo Tree Search (MCTS) \cite{Lauri2016, Patten2018, Best2019}, or global optimization \cite{Popovic2017, Meera2019}. 
While being able to find near-optimal paths over long horizons, 
they suffer from high computational costs
due to repeated predictions of possible future observations. This is particularly exacerbated by computationally expensive visibility checks in obstacle-rich environments.
The resulting long re-planning times make them unpractical for time-constrained and fast-paced dynamic scenarios.
These computational issues are partially addressed by \cite{Schlotfeldt2018, Cao2021}, but as the aforementioned methods, they simplify or do not consider the kinodynamic constraints of the robot.


Local motion planners can ensure dynamic feasibility and collision avoidance, but this might result in a trajectory deviating from the planned informative path. 
Maximizing an information-theoretic objective directly in local trajectory optimization has been proposed for SLAM \cite{Leung2006a, Ryan2010} and grid mapping \cite{Charrow2015}. 
This approach requires a differentiable information gain model and is computationally expensive for long planning horizons.

Our proposed framework, in contrast, combines fast online execution times with non-myopic reasoning and explicit dynamic feasibility and is flexible with respect to observation and information model design choices. This is enabled by combining local motion planning with DRL.

\subsubsection{DRL for Information Gathering}
Thanks to their fast execution times and ability to choose actions conditioned on the recent history of observations, DRL-based approaches have the potential to find a suitable trade-off between quickly reacting to new observations and efficient information gathering.
A common component of previous DRL-based information gathering approaches \cite{Jeong2019, Chen2019, Zhu2018, Niroui2019, Viseras2019, Julian2019, Wei2020, Chaplot2020} is that the agent's current incomplete knowledge about the environment is formulated as an observation to reason about where informative sensor measurements can be taken.
The methods differ in the type of actions being selected, and thus the way the policy interacts with the robot. A common approach is to select from a discrete set of motion primitives \cite{Jeong2019, Chen2019, Julian2019} for a simplified first-order dynamic model, and directly apply them to the robot.
In other works, the learned policy makes higher-level decisions (e.g., by choosing the next frontier \cite{Niroui2019}, subgoal \cite{Chaplot2020}, subregion \cite{Zhu2018} or graph vertex \cite{Wei2020, Viseras2019} to observe). In \cite{Niroui2019, Zhu2018, Chaplot2020}, that action is executed by a lower-level path planner.

However, none of the mentioned works does explicitly account for how the actions chosen by the DRL agent can result in dynamically feasible, collision-free trajectories. 
Our method trains a policy to give a high-level local
subgoal, or \textit{viewpoint} reference, to a lower-level MPC trajectory planner that can ensure the satisfaction of the robot’s constraints. In contrast to the global subgoal policy in \cite{Chaplot2020}, the subgoals in our method are restricted to the robot's local surroundings and continuously guide a dynamics-aware trajectory planner.
%
Similar to \cite{Chen2019}, we use the current knowledge of the global map and local observations as inputs to our policy, but also include the robot's dynamical state to allow for reasoning about the behavior of the underlying MPC. \looseness=-1


\subsection{Contribution}

The main contributions of this work are the following:
\begin{itemize}
   \item An informative trajectory planning hierarchical framework combining a viewpoint recommendation policy with receding-horizon trajectory optimization. Our method plans safe and 
  dynamically
   feasible trajectories while navigating the robot to informative observations. 
   \item A method for training a DRL agent together with a local motion planner, such that the policy learns to guide the motion planner in an obstacle-rich environment and to maximize the cumulative information-theoretic reward.
\end{itemize}

We present simulation results comparing our method with an MCTS planner and a greedy policy, in terms of the execution time and information gathering performance. We aim at significantly faster execution than MCTS, with little loss of performance, and substantially better performance than with the greedy policy.
Additionally, we present qualitative results demonstrating the exploration behavior of our method.


\section{Preliminaries}
\label{sec:3_problem}


\subsection{Problem Formulation}
\label{sec:2_problem}

Consider a robot that has to explore an unknown 2D environment $\mathcal{W} \subset \mathbb{R}^2$ in order to find an unknown number of targets in this environment. 
The dynamics of the robot are described by a discrete-time model $\mathbf{x}_{t+1} = f( \mathbf{x}_t, \mathbf{u}_t ) $
where $\mathbf{x}_t$ is the state of the robot, and $\mathbf{u}_t$ is the control input applied at time step $t$. We assume that $\vx_t$ is observed, e.g., using onboard sensing. 
We denote the position of the robot in $\cW$ at time $t$ by $\vp_t = [x_t,y_t]^T$, $\vp_t \in \mathcal{W}$. The area covered by the robot at time $t$ is denoted by $\cO_t$. The robot must avoid collisions with static obstacles $\mathcal{O}_\text{obst} \subset \mathcal{W}$.

When moving in the environment, the robot builds, from sensor observations, a map
about possible target locations in the environment. We model this target map as a probabilistic occupancy grid map \cite{Thrun2005}, represented by the random variable $\mathbf{M}$ (see \cref{sec:2_prelim_belief}).
The observation vector is modeled as a random variable $Z_t$, with a realization denoted by $z_t$.
At each time step $t$ the robot makes an
observation $Z_t$ about nearby targets at its current position $\vp_t$, and updates its belief about the target map $\vM$.
Subsequently, the control inputs are computed that move the robot to its next observation pose $\mathbf{p}_{t+1}$. \looseness=-1

The objective of the robot is to reduce the uncertainty in the target map $\vM$ by making informative observations $Z_t$. We formalize this objective as maximizing the cumulative mutual information (MI) between the robot's prior about $\vM$ at time step $t$, and the latest measurement $Z_t$, given the history of measurements until the last time step, $z_{0:t-1}$. The MI quantifies the reduction in uncertainty by making observation $Z_t$, and it is denoted by $I(\mathbf{M};Z_{t}|z_{0:t-1})$ \cite{Julian2014}.

The informative trajectory planning problem then is to maximize the cumulative MI while ensuring a collision-free, kinodynamically feasible trajectory over a horizon $L$ (the total time-budget for the mission) and given an initial state $\vx_0$ and an initial observation $z_0$: 
\begin{subequations}
\label{eq:problem}
\begin{align}
    \label{eq:problem_obj}
    \max_{\vu_{0:L-1}} \quad& \sum_{t=1}^L  I(\mathbf{M};Z_{t}|z_{0:t-1})
    \\
    \label{eq:problem_dynamics}
    \text{s.t.} \quad& \mathbf{x}_{t} = f(\mathbf{x}_{t-1}, \mathbf{u}_{t-1})
    \\
    \label{eq:problem_colav}
    & \cO_{t} \cap \mathcal{O}_\text{obst} = \emptyset
    \\
    \label{eq:problem_observation}
    & Z_{t} = h(\vx_{t}),
    \\
    \label{eq:problem_sets}
    & \mathbf{x}_{t} \in \mathcal{X}, \mathbf{u}_{t-1} \in \mathcal{U},\ 
    \\
    & t = 1, ..., L \nonumber
\end{align}
\end{subequations}
where \eqref{eq:problem_dynamics} is the constraint on the robot dynamical model (\cref{sec:2_robot_dynamics}), \eqref{eq:problem_colav} is the collision avoidance constraints, and $\mathcal{X},\mathcal{U}$ are the admissible sets of robot states and control inputs, respectively.
\cref{eq:problem_observation} is the observation model, described in \cref{sec:2_prelim_belief_obs}.

\subsection{Belief Map and Observation Model}
\label{sec:2_prelim_belief}
\subsubsection{Belief Map}
The target map $\mathbf{M}$ is a discretization of $\mathcal{W}$ in $n$ grid cells, associated with independent Bernoulli random variables $M_{i} \in \{ 0,1 \},\ \forall i \in \{ 1,...,n \}$, with 1 indicating target occupancy, and 0 otherwise. 
The robot's \textit{belief} about the map $\vM$ is described by probabilities of target occupancy in each cell $i$, denoted by $P_{t,i} \coloneqq \mathbb{P}(M_{i} = 1|z_{0:t})$, 
and initialized with a uniform prior of $\mathbb{P}(M_{i} = 1)=0.5$.
Given a new observation $z_t$, 
the Bayesian update
of $P_{t,i}$ using log odds \cite{Thrun2005} is: \looseness=-1
\begin{equation}
    \label{eq:problem_inf_update}
    l(M_{i} | z_{0:t}) = l(M_{i} | z_{0:t-1}) + l(M_{i} | z_{t}),
\end{equation}
where $l(M_{i} | z_{t})$ is an inverse sensor model \cite{Thrun2005}.

\subsubsection{Observation Model}
\label{sec:2_prelim_belief_obs}
To make observations $Z_t$, the robot is equipped with
a sensor that can detect targets up to a distance $d_{\max}$ from the robot and within a field-of-view of \SI{360}{\degree} and associate it with a cell in the map $\mathbf{M}$.
The set of cells visible from position $\vp_t$ is denoted by $\cI_t$. It only includes cells for which the visibility of its center point is not occluded by obstacles $\cO_{obst}$.
The observation made at each time step is a vector composed of the individual cell observations of target occupancy, namely $Z_t \in \{0,1\}^{|\cI_t|}$.
\cref{eq:problem_inf_update} is only applied to the cells in $\mathcal{I}_t$ after each observation. 

The mutual information between the prior about the target map $\vM$ and an observation $Z_t$ is equal to the reduction of the conditional entropy in $\vM$ by observation $Z_t$ \cite{Cover2005, Julian2014}:
\begin{align}
    I(\mathbf{M};Z_{t}|z_{0:t-1}) &= H(\vM|z_{0:t-1}) - H(\vM|Z_t,z_{0:t-1}) 
\end{align}
where $H(\vM)=\sum^{n}_{i=1}H(M_{i})$ is the entropy of the target map and $H(M_{i})$ are the respective cell entropies.

\subsection{Robot Dynamics}
\label{sec:2_robot_dynamics}

We consider the robot to be modeled by a second-order
unicycle model
\cite{lavalle2006planning}
\begin{equation}\label{eq:dynamics_model}
    \begin{array}{lll}
    \dot{x}=v\cos{\psi} & \dot{v}=u_a & \dot{\psi}=\omega \\
    \dot{y}=v\sin{\psi} & \dot{\omega}=u_\alpha . & \\
    \end{array}
\end{equation}
which is discretized with sampling time $T_S$.
Thus the state of the robot is described by $\mathbf{x}=[x, y, \psi, v, \omega]^T$, where $\psi$ the heading angle in a global frame, $v$ denotes the robot's longitudinal velocity, and $\omega=\dot{\psi}$ the angular velocity. The control input $\mathbf{u} = [u_a, u_\alpha]^T$ consists of the robot's linear and angular acceleration, respectively.

\section{Method}
\label{sec:3_method}

We hierarchically solve the problem in \eqref{eq:problem} by separating it into a high-level sequential decision-making problem and a local trajectory planning problem. The first aims at determining a reference viewpoint, such that future information gains are maximized based on the current belief following from past observations (\cref{sec:3_method_rl}). The local trajectory planning problem aims at moving the robot towards the recommended viewpoint while ensuring kinodynamic feasibility and collision avoidance (\cref{sec:3_method_mpc}). The concept of the proposed framework is depicted in \cref{fig:overview}. Our proposed approach builds on \cite{Brito2021}, extending its task and environment scope for information gathering in obstacle-rich~environments. 

\subsection{Reinforcement Learning of Viewpoint Recommendations}
\label{sec:3_method_rl}

Our method learns, via reinforcement learning, a policy $\pi$ that recommends 
every $N_a$ timesteps
a reference position $\mathbf{p}_t^{{\textrm{ref}}}$ in the robot's neighborhood (the reference viewpoint) to an MPC motion planner, such that the resulting trajectories of the robot lead to observations that maximize rewards, and eventually result in near-complete coverage of the available information in the environment.

\subsubsection{Observation}
\begin{figure}[t]
    \centering
    \includegraphics[width=0.9\columnwidth, trim=60 0 120 0]{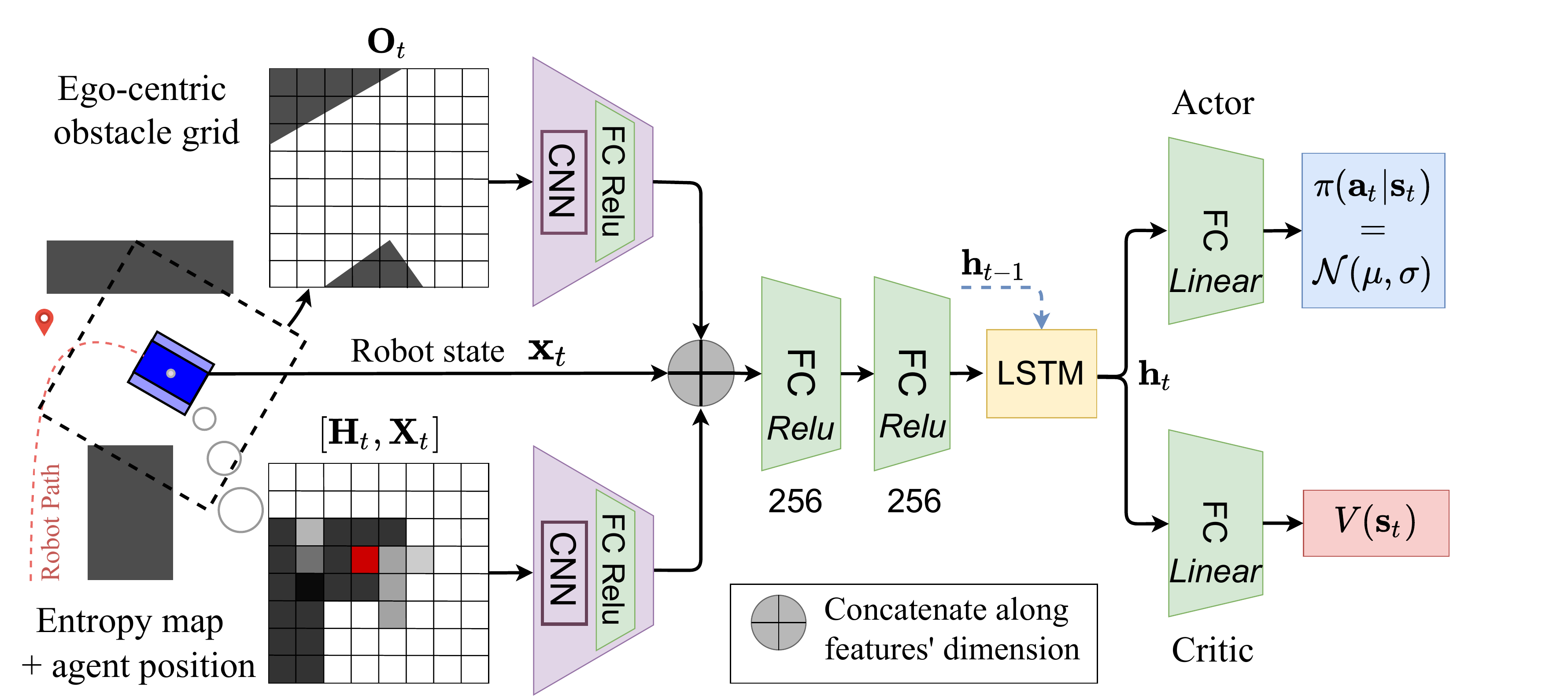}
    \caption{Proposed policy and value function network, with two encoders processing an ego-centric obstacle grid map $\vO_t$ and a two-channel image representing the belief $[\mathbf{H}_t, \vX_t]$, respectively. The entropy map $\mathbf{H}_t$ is depicted as a grayscale image, with darker shades corresponding to lower uncertainties.
    The second channel $\vX_t$ is visualized by a red grid cell marking the current agent position. The encoder structure and hyperparameters are as in \cite{Pfeiffer2018}, $\mathbf{h}$~is the LSTM hidden state, and FC refers to a fully-connected layer. \looseness=-1}
    \label{fig:network}
    \vspace{-2ex}
\end{figure}

The goal is to learn a policy that uses the robot's own belief about $\mathbf{M}$, and local observations about nearby obstacles 
$\mathbf{O}_{t} \in \mathcal{O}_\text{obst}$. Both inputs are visualized in \cref{fig:network}.
The local obstacle observation $\mathbf{O}_{t}$ is a binary grid map of obstacles around the robot, given as an $m \times m$ image, centered at the robot's position and aligned with its orientation \cite{Pfeiffer2018, Chen2019}.
Such an egocentric observation improves generalization across different environments.
The robot's belief about $\vM$ is represented by a map of cell entropies $H(M_i)$, denoted by $\mathbf{H}_t$, that informs the agent about uncertainties in different map regions. 
An indicator function map $\mathbf{X}_t$ for the agent's position in the map is appended as a second channel to $\mathbf{H}_t$ \cite{Zhu2018, Niroui2019}.
Hence, at time step $t$ the RL observation vector $\mathbf{s}_t$ \looseness=-1
is
\begin{equation}
    \mathbf{s}_t
    = \left[ \mathbf{H}_t, \mathbf{X}_t, \mathbf{O}_t, \mathbf{x}_t \right]^{\textrm{T}},
\end{equation}
where $\mathbf{x}_t$ is the robot's state defined in \cref{sec:2_robot_dynamics}. 

\subsubsection{Action}
The RL action $\mathbf{a}_t \in \mathcal{A} \subset \mathbb{R}^2$ is defined as the relative position
$\delta_t$ of the viewpoint reference with respect 
to the robot's current position,
\begin{equation}
\begin{aligned}
    \mathbf{a}_t &= \delta_t \sim \pi(\va_t|\mathbf{s}_t)
    \\
    \mathbf{p}_{t}^\textrm{ref} &= \mathbf{p}_t + \delta_t
\end{aligned}
\end{equation} 
The position increment is constrained inside a square around the robot, such that the continuous action space of our RL method is $\cA = \{\ \delta_t \in \R^2 \ |\ \norm{\delta_t}_\infty \leq \delta_{\text{max}} \}$.

\subsubsection{Reward Function}
The main objective of the informative trajectory planning problem \eqref{eq:problem} is to maximize the information gains. Hence, we define an information-theoretic reward function using the mutual information gained through the observation $Z_{t+N_a}$, made $N_a$ steps after the last action $\va_t$.
Moreover, we add a term $r_\textrm{pen}$ penalizing each time step, incentivizing the agent to achieve the coverage goal, terminating the episode, as soon as possible.
The reward function is defined as \looseness=-1
\begin{equation}
    r(\mathbf{s}_t,\mathbf{a}_t) = 
    I(\vM;Z_{t+N_a}|z_{0:t}) 
    + r_\textrm{pen} .
\end{equation}

\subsubsection{Policy Network Architecture}

\cref{fig:network} depicts our proposed policy network architecture.
We employ two CNN models using the architecture and hyperparameters proposed in \cite{Pfeiffer2018} to encode spatial information in the two image inputs $[\mathbf{H}_t, \mathbf{X}_t]$ and $\mathbf{O}_t$.
These encoder networks are each trained by gradients coming from both the policy update loss (\cref{sec:3_method_training})
and a reconstruction loss generated using a decoder network reversing the encoder operations \cite{Pfeiffer2018}.
Thus, compressed latent representations of spatial features in the local obstacle grid and the entropy map are learned, which the policy can exploit to learn actions that guide the robot around nearby obstacles and to map regions with high uncertainties.
The two latent feature vectors are concatenated with the state $\mathbf{x}_t$ of the robot's dynamical model, so that the policy can learn how to guide the MPC planner using viewpoint references with respect to its closed-loop dynamical behavior.

After feeding the full feature vector into two fully-connected (FC) layers, an LSTM layer \cite{Hochreiter1997} models the time-dependencies between previous states and the current state.
The hidden state of the LSTM is fed to the final actor and critic heads modeled as FC linear layers. 
We model the policy $\pi$ as a diagonal Gaussian distribution, i.e. $\pi_{\theta}(\va_t|\mathbf{s}_t) = \mathcal{N}(\mu, \sigma^2)$, such that
$\delta_t \sim \pi_\theta(\va_t|\mathbf{s}_t)$. The distribution's mean $\mu$ and log-standard-deviation $\log \sigma$ are learned by the actor head.
The critic head estimates the state-value function $V^\pi_\theta (\vs_t) = \mathbb{E}_\pi \left[ \sum^\infty_{t=0} \gamma^t r(\mathbf{s}_t,\mathbf{a}_t) \right]$ of the current policy, where $\gamma$ is the discount factor. 
The subscript $\theta$ denotes the current network~parameters. 

\subsection{Local Planner}
\label{sec:3_method_mpc}
We rely on receding-horizon trajectory optimization to generate control commands for the robot satisfying dynamic and collision constraints. For dynamic constraints we employ a second-order unicycle model as defined in \cref{sec:2_robot_dynamics}. 

For the collision constraints, we assume the robot's space, $\mathcal{O}_t$, to be a circle with radius $r$, and  each obstacle's space is defined as a polygon.
To ensure collision-free motions, first, we compute a linear constraint to ensure that the robot's space does not overlap with static obstacle's space, i.e., $\pazocal{O}_t(\vx_t) \cap \pazocal{O}_\text{obst}  = \emptyset$, at planning step $k$ defined as 
\begin{equation} \label{eq:constraints}
    c^{o_j}_k = {\vn_k^{o_j}}^{\textrm{T}}\vp_k \leq b_j - r,
\end{equation}
where $\vn_k^{o_j}$ is the normal vector at the closest point $\vp^{o_j}_k$
on the surface of the $j$\-/th obstacle
and $b_j = -{\vn_k^{o_j}}^{\textrm{T}}\vp^{o_j}_k$. 
To limit the complexity of the optimization problem, we only consider a set of $n_{obs}$ constraints for the closest obstacles. The distance between the robot's position and the $j$-th linear constraint is computed as:
\begin{equation} \label{eq:distance}
    \norm{\vp_k,c_k^{o_j}} = \frac{\abs{{\vn_k^{o_j}}^{\textrm{T}}\vp_k+b_j}}{\norm{\vn_k^{o_j}}}
\end{equation}

The DRL policy provides a reference position $\vp^{\textrm{ref}}_t$ (viewpoint) guiding the robot to maximize future rewards. Similarly to \cite{Brito2021}, we define a terminal cost enabling the robot to reach the provided viewpoint reference: 
\begin{align}
    J_N(\vx_{t+N},\vp^{\textrm{ref}}_t) =  \norm{\frac{\vp_{t+N} - \vp^{\textrm{ref}}_t}{\vp_{t} - \vp^{\textrm{ref}}_t}}_{Q_N},
\end{align}
where $\vp_{t+N}$ is the robot's terminal position (at planning step $N$) and $Q_N=\text{diag}(q_N, q_N)$ is the terminal cost matrix.
To generate smooth trajectories, we employ a quadratic penalty on the control commands as a stage cost for planning step $k$:
\begin{align}
    J_{k}^{\vu}(\vu_k) = \norm{\vu_k}_{Q_{u}}
\end{align}
where $Q_{u}=\text{diag}(q_a, q_\alpha)$ is the stage cost matrix.

At every time step $t$, a non-convex optimization problem is solved with planning horizon $N$ under the kinodynamic and collision constraints, given the initial state $\vx_t \in \cX$: 
\begin{align}
    \label{eq:control_problem}
    \min_{\vu_{t:t+N-1}}  ~        
                            & \sum_{k=t}^{t+N-1} J^{u}_k(\vu_k) 
                          +  J_N(\vx_{t+N},\vp^{\textrm{ref}}_t)
                            \nonumber \\
    \text{s.t.}	~~	    &  \mathbf{x}_{k+1} = f(\mathbf{x}_k, \mathbf{u}_k), \\
                            & c_{k+1}^{o_j} \leq b_j - r,\ \forall j \in \{1,\dots,n_{\textrm{obs}}\},
                            \nonumber \\
                            & \vu_{k} \in \cU,\ \vx_{k+1} \in \cX,\
                            \forall k \in \{t,...,t+N-1\}.
                            \nonumber
\end{align}
The equality constraint is the discrete-time model of the continuous dynamics model presented in \eqref{eq:dynamics_model}.

\subsection{Training Procedure}
\label{sec:3_method_training}

First, we warm-start the policy training with behavior cloning updates, using the one-step greedy policy outlined in \cref{sec:4_results_baseline_greedy}, which outputs $\vp^{\textrm{ref}}_t$, in combination with MPC \eqref{eq:control_problem} as the expert policy. We define the 
expert reference viewpoint as $\va_t^* = \vp_N^* - \vp_t$, where $\vp_N^*$ is 
the last
position in the MPC-generated trajectory. 
\pagebreak \noindent
For the first $N_{SL}$ policy steps of the training, we apply $\va_t^*$ as the agent's action and use it as a label to perform supervised learning of the policy $\pi_\theta$, as described in \cite{Brito2021}.
Subsequently, the policy is trained with DRL using Proximal Policy Optimization (PPO) \cite{Schulman2017} 
until reaching $N_\text{train}$ policy steps.
One policy step yielding a new viewpoint corresponds to $N_a$ timesteps, and 
MPC is executed at every time step $t$ with the last sampled viewpoint reference. 
The PPO horizon is $S=L/N_a$ policy steps.

\begin{figure}
    \centering
    \hspace{1.2mm}
    \begin{subfigure}{0.18\columnwidth}
        \includegraphics[width=\textwidth, trim= 200 50 80 50, clip]{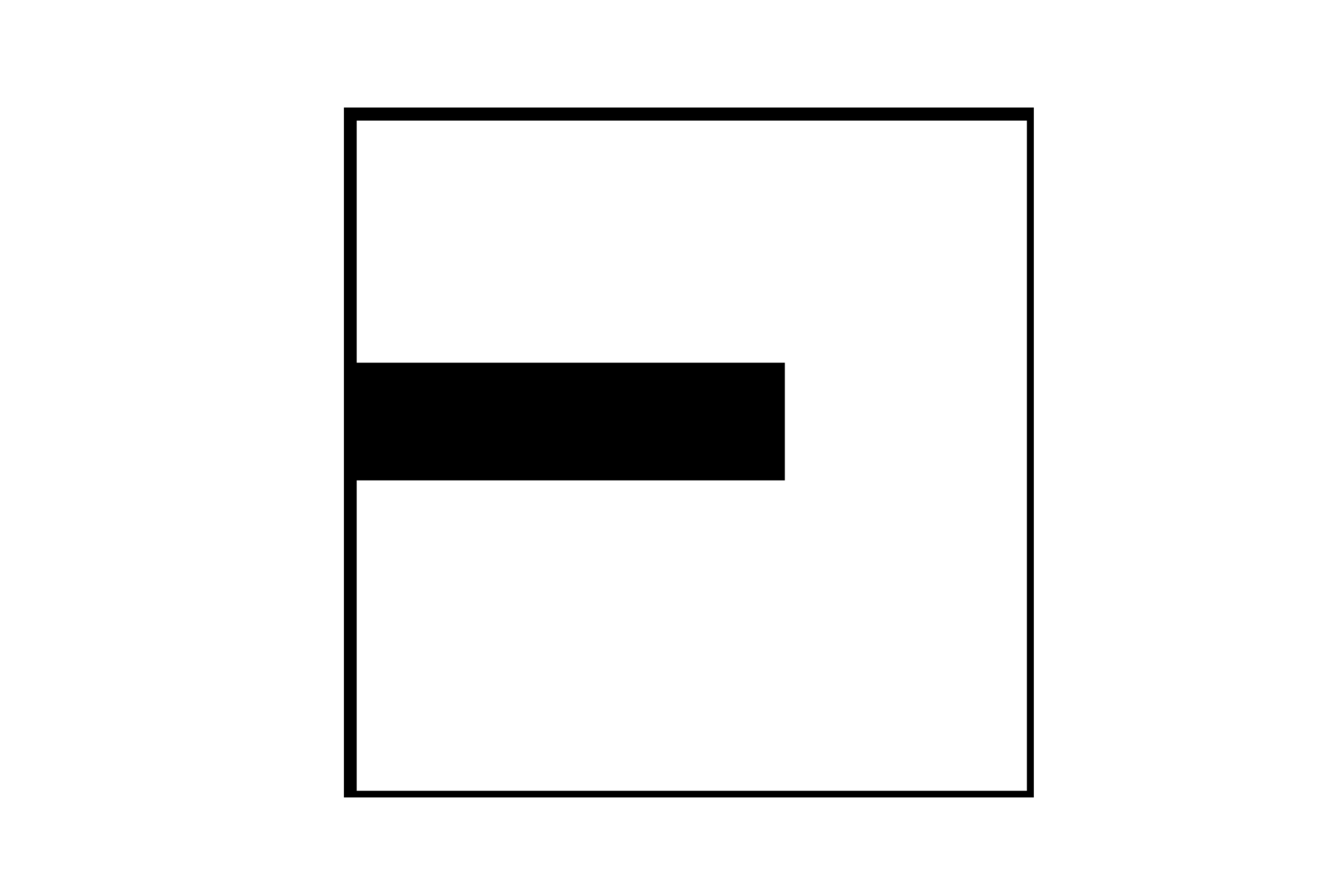}
    \end{subfigure}
    \hspace{-4mm}
    \begin{subfigure}{0.18\columnwidth}
        \includegraphics[width=\textwidth, trim= 200 20 80 20, clip]{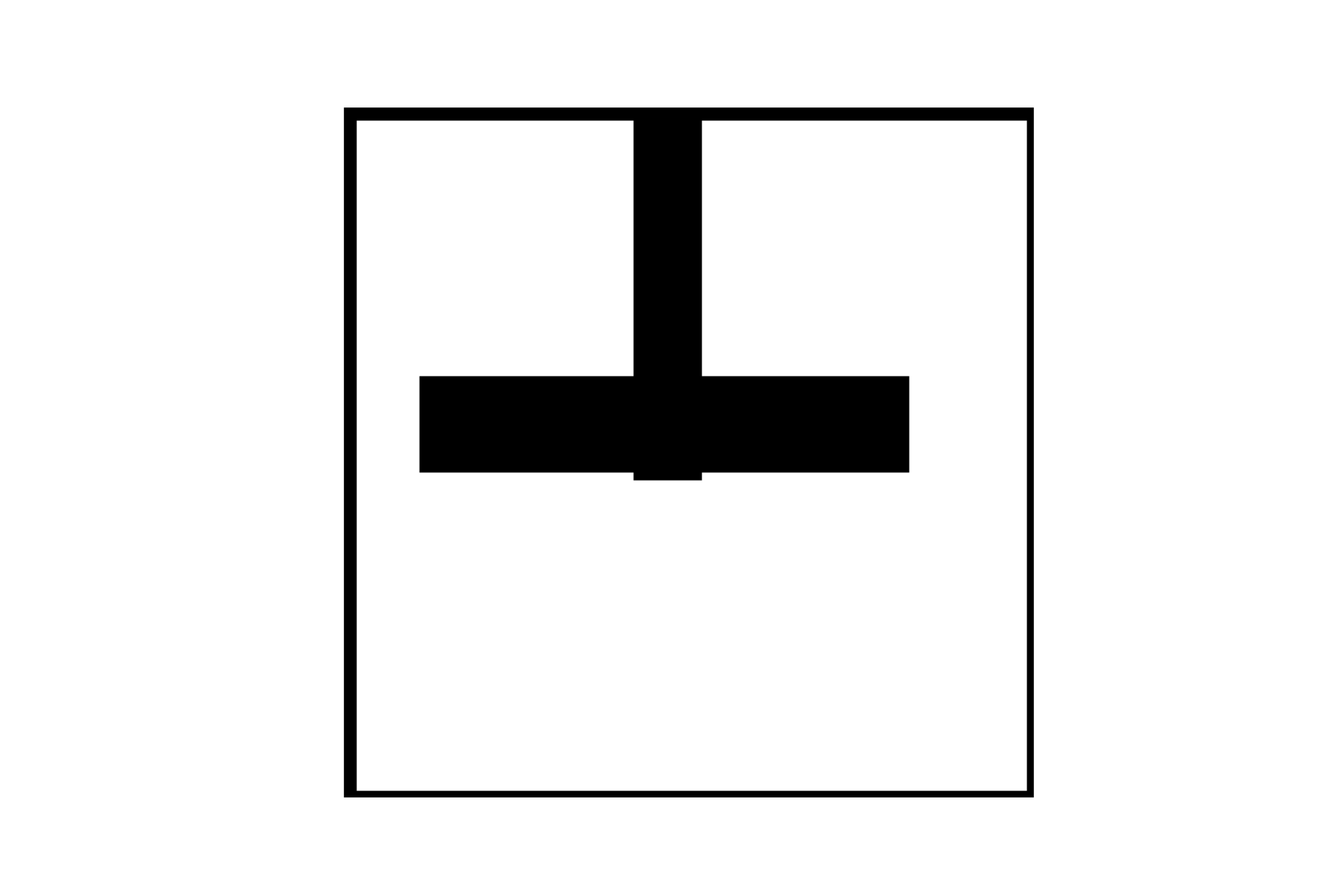}
    \end{subfigure}
    \hspace{-4mm}
    \begin{subfigure}{0.18\columnwidth}
        \includegraphics[width=\textwidth, trim= 200 20 80 20, clip]{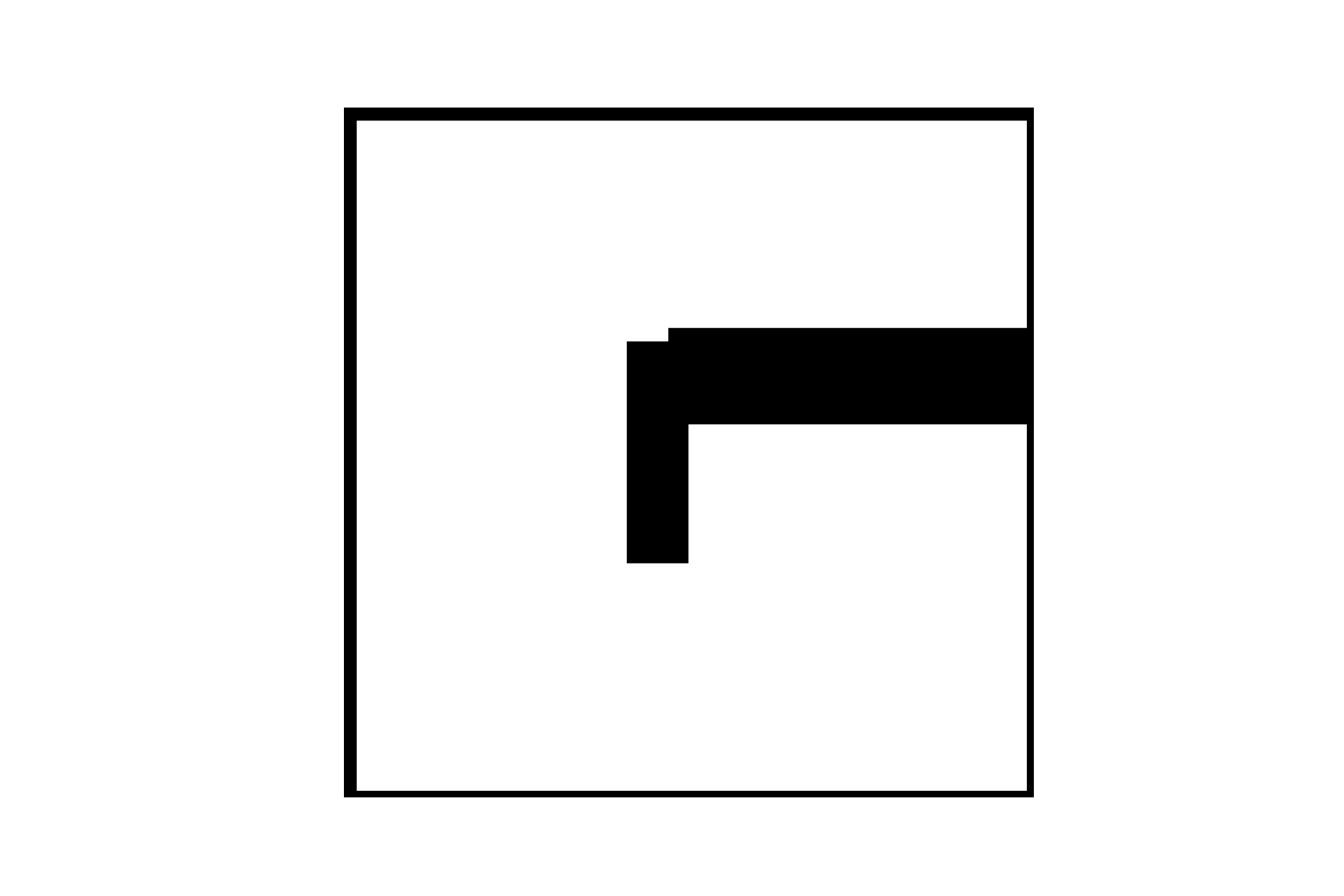}
    \end{subfigure}
    \hspace{-4mm}
    \begin{subfigure}{0.18\columnwidth}
        \includegraphics[width=\textwidth, trim= 200 20 80 20, clip]{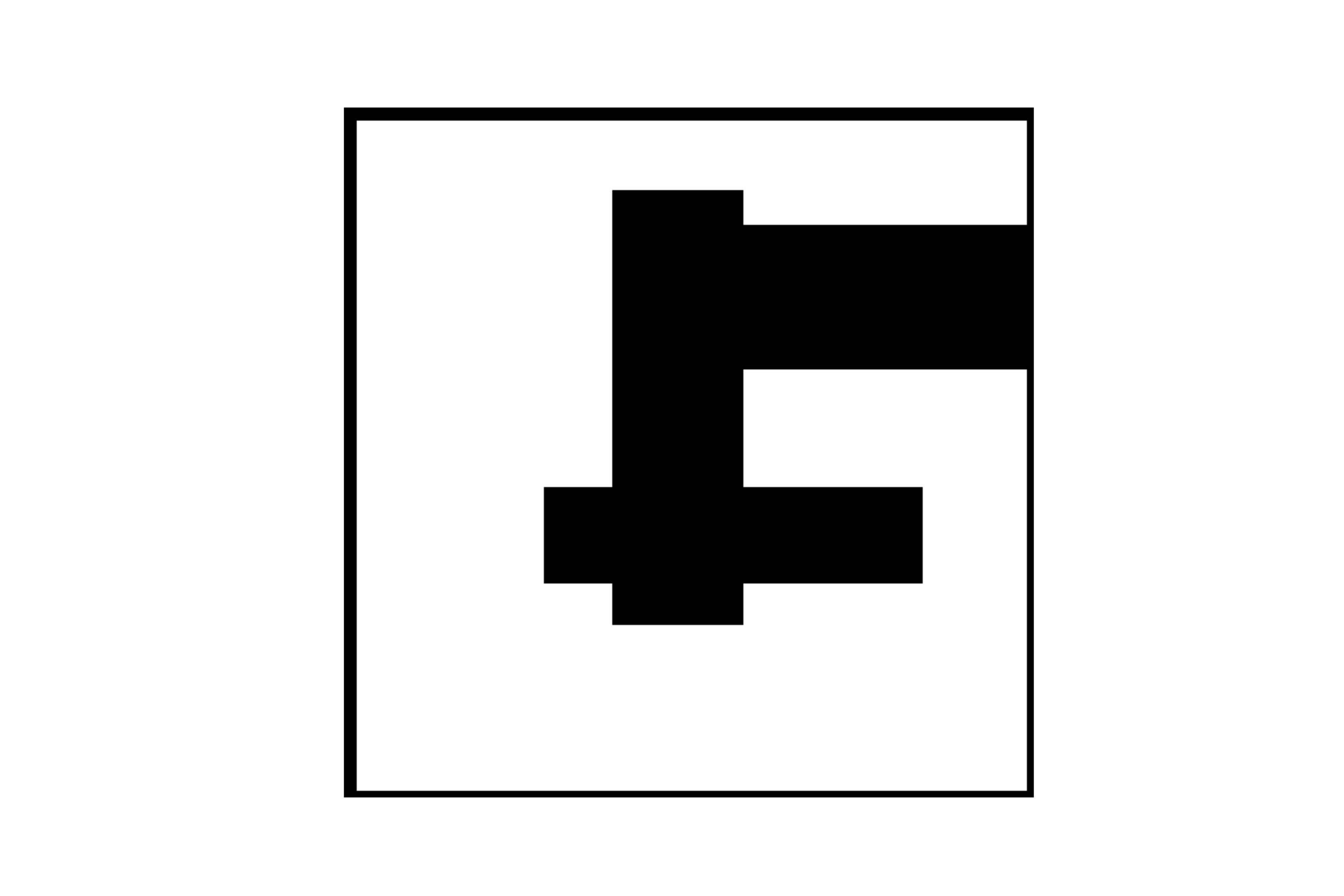}
    \end{subfigure}
    \hspace{-4mm}
    \begin{subfigure}{0.18\columnwidth}
        \includegraphics[width=\textwidth, trim= 200 20 80 20, clip]{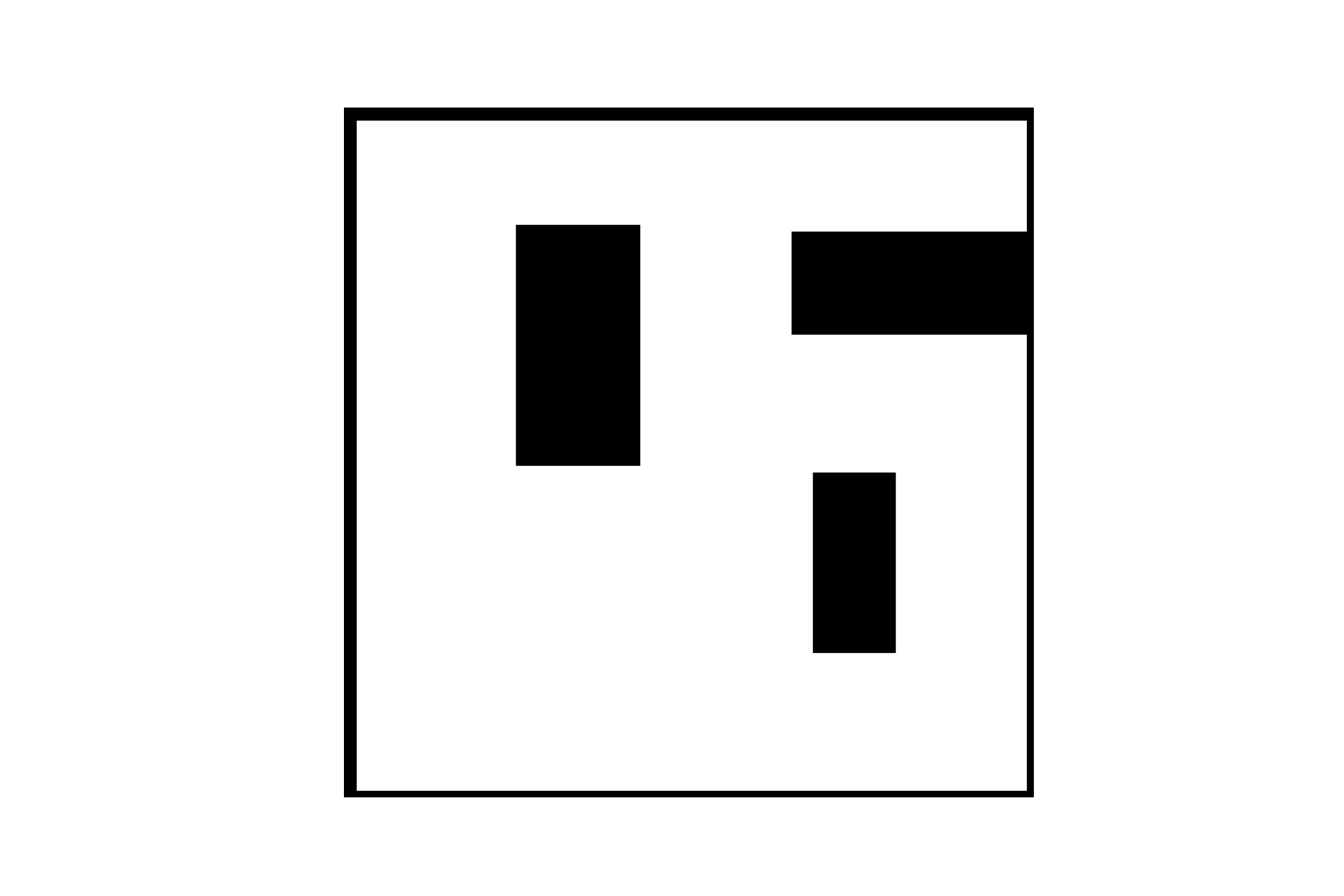}
    \end{subfigure}
    \hspace{-4mm}
    \begin{subfigure}{0.18\columnwidth}
        \includegraphics[width=\textwidth, trim= 200 20 80 20, clip]{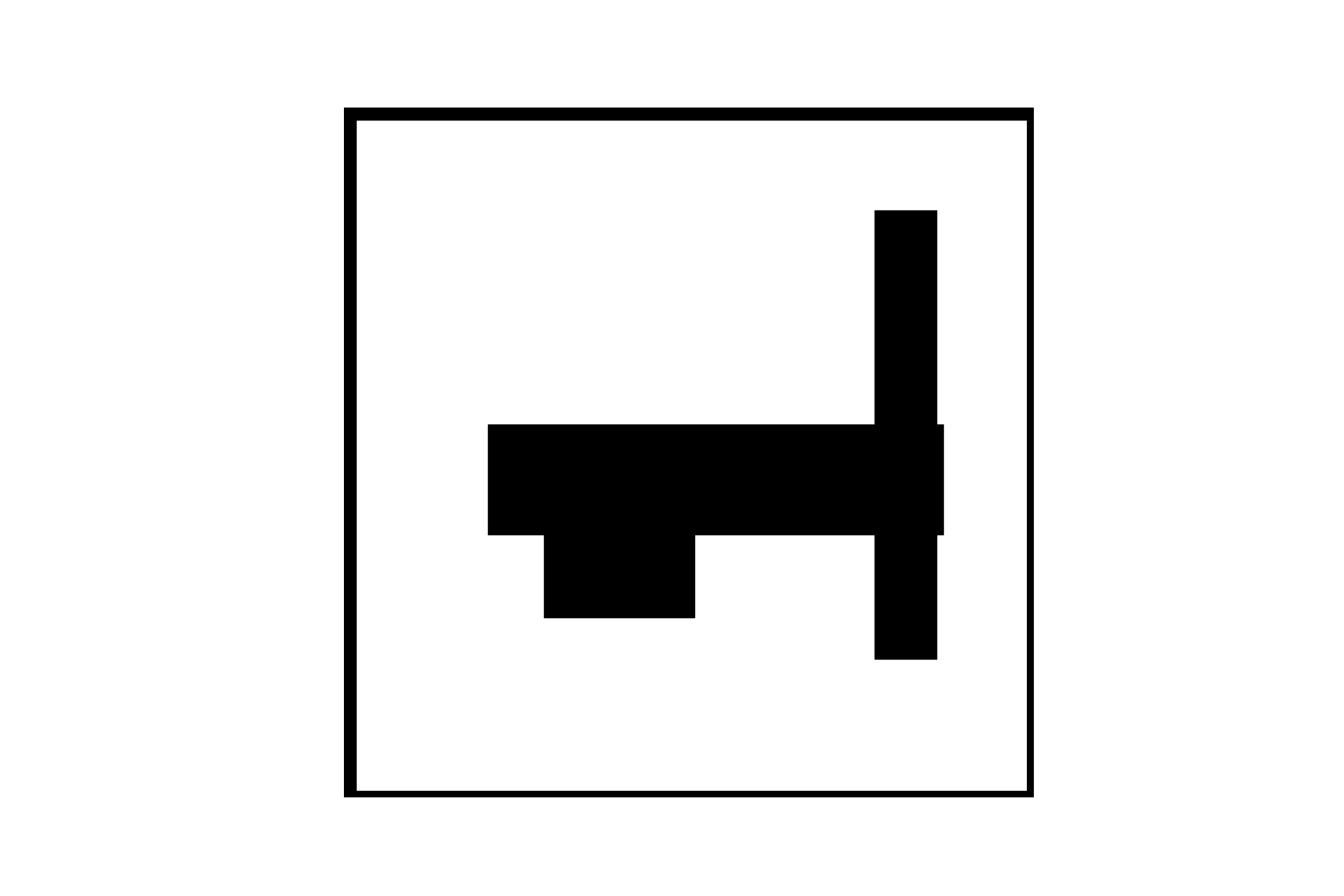}
    \end{subfigure}
    \caption{Examples of the random environments used in training.}
    \label{fig:example_obstacles}
    \vspace{-2ex}
\end{figure}

Training and testing are performed in randomly generated environments as depicted in \cref{fig:example_obstacles}, with the agent initialized at a random position. Random rectangular obstacles are generated and environments, where obstacles block the agent from reaching the entire free space, are omitted. We employ curriculum learning during training \cite{Bengio2009}, increasing the number of obstacles from one to three.


Episodes are terminated after the completion of the information gathering task, or if a maximum number of time steps $t_\text{max}$ is reached (\textit{failure}). 
The task is completed when, at time $t$, the conditional entropy of the belief about the map cells in the free space $\cW \setminus \cO_\text{obst}$, denoted by $\vM^{\text{free}}$,
is smaller than a predefined ratio of the  entropy of the initial belief prior.
That is, when
$H(\vM^{\text{free}}|z_{0:t}) \leq (1-\beta) H(\vM^{\text{free}})$, 
where $\beta \in [0,1)$ is the share of information that should be gathered by the robot, defining the coverage goal.

\section{Results}
\label{sec:4_results}

In this section, we present quantitative and qualitative simulation results of the proposed method. We compare the performance metrics of our method against two baseline approaches (\cref{sec:4_results_perf}) and analyze the behavior of the informative trajectory planning method (\cref{sec:4_results_qa}). The baselines are introduced in \cref{sec:4_results_baselines}, and the simulation setup for training and testing is outlined in \cref{sec:4_results_setup}.

\subsection{Simulation Setup}
\label{sec:4_results_setup}

The training procedure described in \cref{sec:3_method_training} builds upon the open-source PPO2 implementation provided by the Stable Baselines framework \cite{stable-baselines}. The nonlinear optimization problem \eqref{eq:control_problem} is solved using the Forces Pro solver \cite{FORCESPro}. Simulations are run in the environment provided by the open-source "gym-collision-avoidance" package \cite{Everett2018}. We train the policy $\pi_\theta$ with $N_\text{proc}$ processes for rollouts on a desktop computer with an AMD Ryzen 9 CPU and 64 GB of RAM. \cref{tab:hyperparameters} presents the hyperparameters used.

\renewcommand{\arraystretch}{1.05}
\begin{table}[t]
\vspace{2.5ex}
\caption{Hyperparameters.}
\vspace{-0.5ex}
\label{tab:hyperparameters}
\centering
\begin{tabular}{|c|c||c|c||c|c|}
\hline
\multicolumn{6}{|c|}{\textbf{MPC}} 
\\ \hline
Horizon $N$ & 15 & 
$T_S$ & \SI{0.1}{\second}  &
$q_N$ & 5.0 
\\ \hline
$q_a$ & \num{0.003} &
$q_\alpha$ & 0.003 &
--&--
\\ \hline
\multicolumn{6}{|c|}{\textbf{Training}} 
\\ \hline
Horizon $S$ & 128 &
$N_a$ & 5 &
$N_\text{proc}$&16
\\ \hline
Learning rate & $10^{-4}$ & 
$\gamma$ & 0.99 &
$N_\text{epochs}$ & 2
\\ \hline
Clip range & 0.2 &
$N_\text{train}$ & $2 \cdot 10^7$ &
$N_{SL}$ & $10^6$ 
\\ \hline
$\delta_{\text{max}}$ & \SI{4}{\meter} &
$t_{max}$ & 640 &
$r_{pen}$ & -0.1
\\ \hline
\multicolumn{6}{|c|}{\textbf{MCTS Baseline} \cite{Best2019}} \\ \hline
$N_\text{tree}$ & 100 &
$N_\text{sim}$ & 10 &
$H_\text{MCTS}$ & 4
\\ \hline
$T_{mp}$ & \SI{1.2}{\second} &
$u_v$ & $\{0, 1, 3\}$ &
$C_\text{UCB}$ & 2.0
\\ \hline
$u_\omega$ & 
\multicolumn{3}{c||}{
$\{-\nicefrac{\pi}{4}, -\nicefrac{\pi}{10}, 0, \nicefrac{\pi}{10}, \nicefrac{\pi}{4}\}$
} &--&-- \\ \hline
\end{tabular}
\vspace{-3.5ex}
\end{table}

\subsection{Baselines}
\label{sec:4_results_baselines}

\renewcommand{\arraystretch}{1.0}

\begin{table*}[t]
\vspace{2.5ex}
 \caption{Performance results, aggregated over 100 random maps with $n \in \{1,2,3\}$ obstacles (see \cref{sec:4_results_perf} for details).}
 \vspace{-0.5ex}
        \centering
\begin{tabular}{|p{2cm}|p{1.0cm}|p{1.0cm}|p{1.0cm}|p{1.0cm}|p{1.0cm}|p{1.0cm}|p{1.0cm}|p{1.0cm}|p{1.0cm}|p{1.0cm}|p{1.0cm}|p{1.0cm}|}
\hline
\multicolumn{2}{|c|}{} & \multicolumn{3}{c|}{Avg. episode rewards (mean ± std)} & \multicolumn{3}{c|}{\% failure } & \multicolumn{3}{c|}{Time until completion [s]}
& \multicolumn{1}{|c|}{Avg. Runtime [s]}
\\ \hline
\multicolumn{2}{|c|}{\# obstacles}                    & \multicolumn{1}{c|}{1}                & \multicolumn{1}{c|}{2}       & \multicolumn{1}{c|}{3}                        & \multicolumn{1}{c|}{1}                &\multicolumn{1}{c|}{2}     & \multicolumn{1}{c|}{3}                       & \multicolumn{1}{c|}{1}                & \multicolumn{1}{c|}{2}       &\multicolumn{1}{c|}{3}  &\multicolumn{1}{|c|}{-}  
\\  \hline \hline
\multicolumn{2}{|l|}{MCTS}                      &  \multicolumn{1}{c|}{19.60 $\pm$ 1.99}     &  \multicolumn{1}{c|}{18.24 $\pm$ 1.09}       &    \multicolumn{1}{c|}{16.79 $\pm$ 2.00}         & \multicolumn{1}{c|}{1}    &  \multicolumn{1}{c|}{2}        &  \multicolumn{1}{c|}{2}        &  \multicolumn{1}{c|}{46.8}   &   \multicolumn{1}{c|}{50.7}       & \multicolumn{1}{c|}{51.7} &  \multicolumn{1}{|c|}{2.486}
\\ \hline
\multicolumn{2}{|l|}{Greedy}                &   \multicolumn{1}{c|}{18.98 $\pm$ 2.25}              &   \multicolumn{1}{c|}{17.50 $\pm$ 2.65}            &  \multicolumn{1}{c|}{15.64 $\pm$ 3.75}       & \multicolumn{1}{c|}{6}    &  \multicolumn{1}{c|}{2}        & \multicolumn{1}{c|}{8}      & \multicolumn{1}{c|}{56.7}                &\multicolumn{1}{c|}{61.6}    &      \multicolumn{1}{c|}{65.7} 
&  \multicolumn{1}{|c|}{0.046}
\\ \hline
\multicolumn{2}{|l|}{Viewpoint Policy (ours)}                            & \multicolumn{1}{c|}{19.41 $\pm$ 0.89}      &  \multicolumn{1}{c|}{18.03 $\pm$ 1.17}          &     \multicolumn{1}{c|}{16.49 $\pm$ 1.41}          & \multicolumn{1}{c|}{0}       & \multicolumn{1}{c|}{1}     &   \multicolumn{1}{c|}{1}            &  \multicolumn{1}{c|}{53.6}                &   \multicolumn{1}{c|}{55.8}       & \multicolumn{1}{c|}{59.6} &   \multicolumn{1}{|c|}{0.004} 
\\ \hline
\end{tabular}
 \label{tab:performance}
 \vspace{-1ex}
\end{table*}

\begin{figure*}
    \centering
    \begin{subfigure}{0.27\textwidth}
        \centering
    	\includegraphics[width=\linewidth, trim = 180 65 0 65, clip]{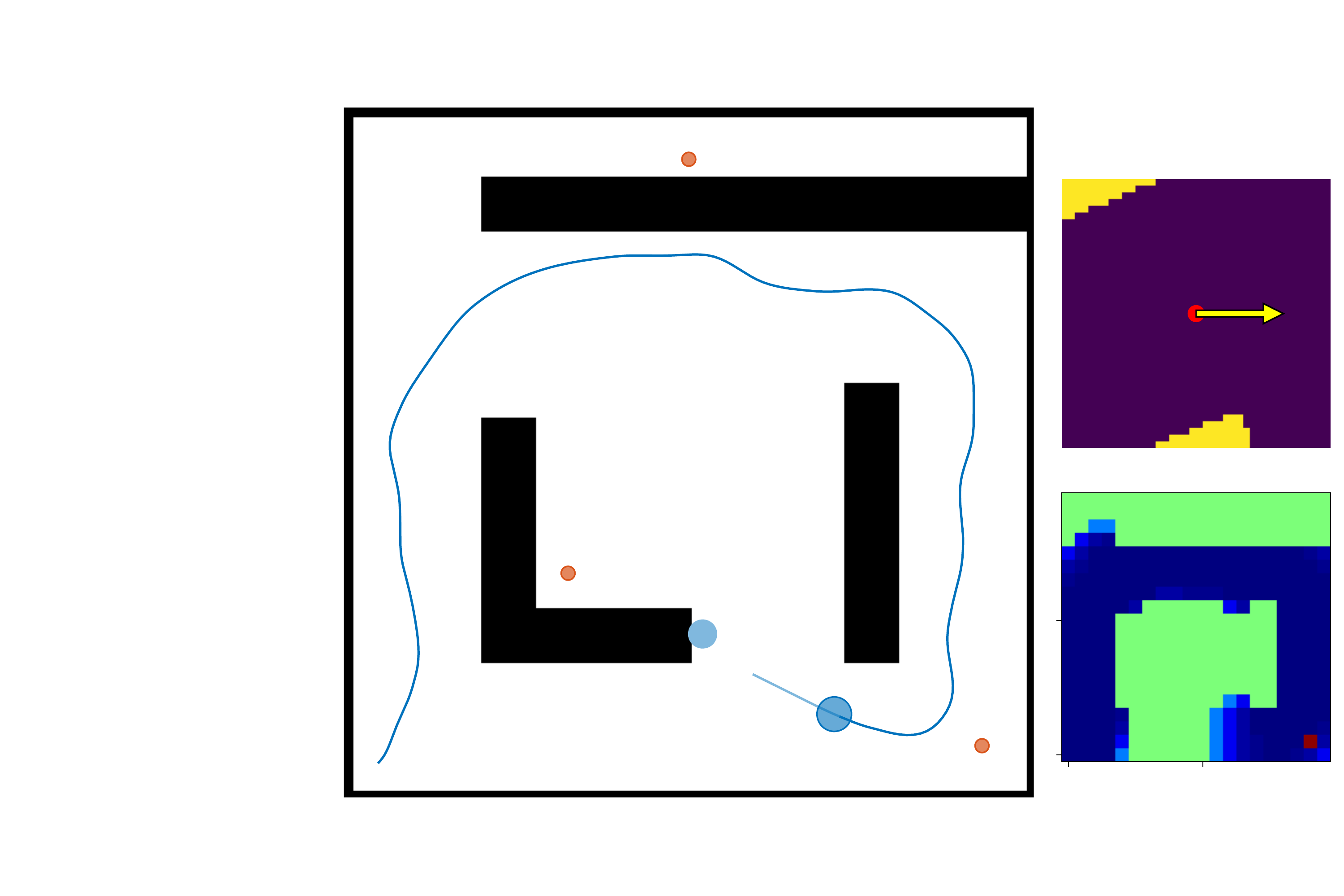}
    	\captionsetup{justification=centering}
    	\caption{$t=\SI{22.5}{\second}$}
    	\label{fig:results_qual_ust1}
    \end{subfigure}%
    \hspace{0.03\textwidth}
    \begin{subfigure}{0.27\textwidth}
        \centering
    	\includegraphics[width=\linewidth, trim = 180 65 0 65, clip]{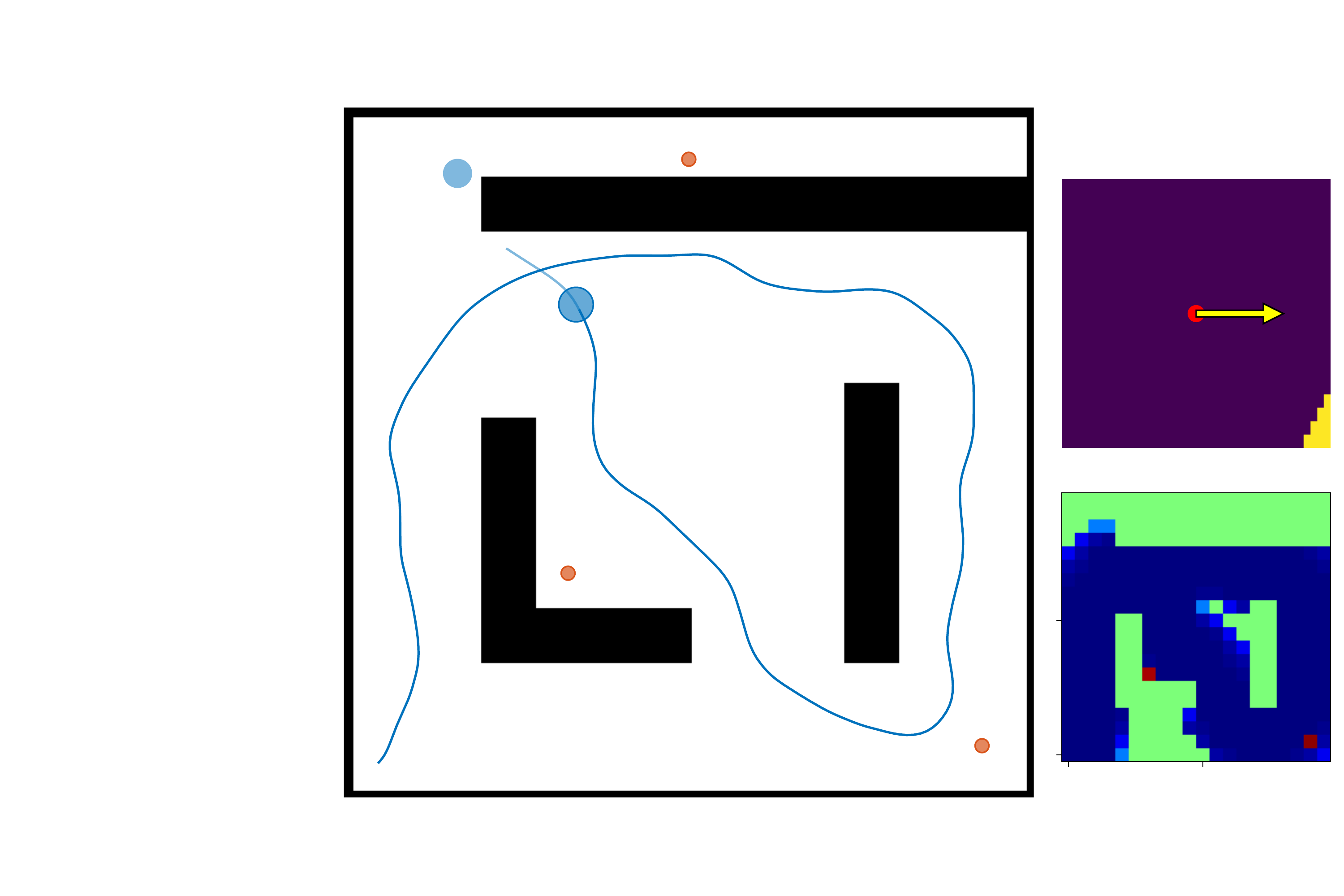}
    	\captionsetup{justification=centering}
    	\caption{$t=\SI{30}{\second}$}
    	\label{fig:results_qual_ust2}
    \end{subfigure}%
    \hspace{0.03\textwidth}
    \begin{subfigure}{0.27\textwidth}
        \centering
    	\includegraphics[width=\linewidth, trim = 180 65 0 65, clip]{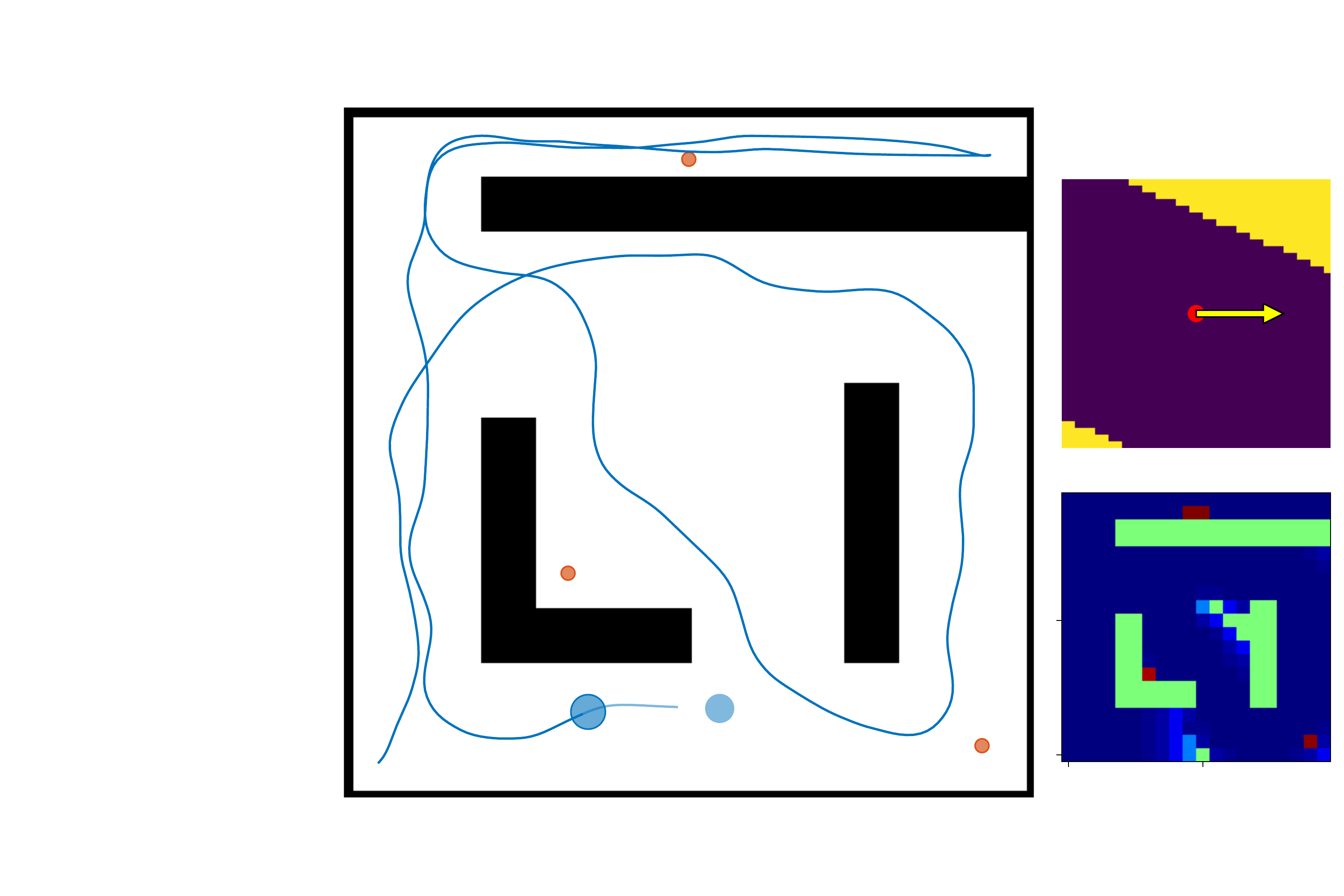}
    	\captionsetup{justification=centering}
    	\caption{$t=\SI{71}{\second}$}
    	\label{fig:results_qual_ust3}
    \end{subfigure}%
    \vspace{0.2ex}
    \caption{Trained policy behavior in an \textit{unstructured} environment of higher complexity than in training, with three timesteps of an episode displayed. The agent effectively explores all areas of the environment and manages to enter and leave the narrow dead-end corridor. The upper-right grid next to each map shows the ego-centric local obstacle observation $\mathbf{O}_t$ of the agent, and the lower-right grid the belief map of the probabilities $P_{t,i}$ of the current belief (\cref{sec:2_prelim_belief}). The colors in the belief map range from dark blue $P_{t,i}=0$ to dark red $P_{t,i}=1$, with the the green areas indicating $P_{t,i}=0.5$ (the initial value). \looseness=-1}
    \label{fig:results_qual_unstr}
\end{figure*}

\begin{figure*}
    \centering
    \begin{subfigure}{0.27\textwidth}
        \centering
    	\includegraphics[width=\linewidth, trim = 180 65 0 65, clip]{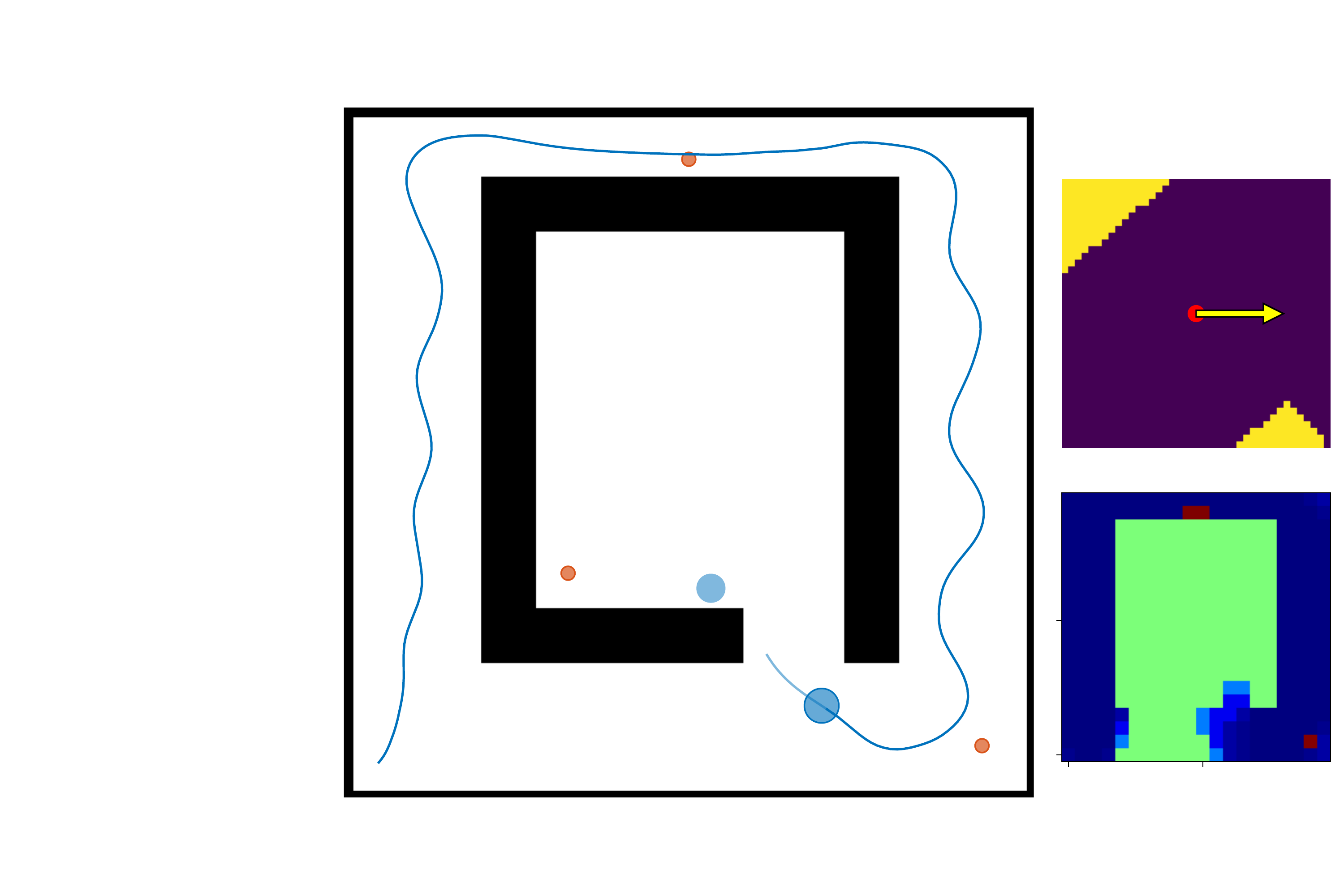}
    	\captionsetup{justification=centering}
    	\caption{$t=\SI{30.5}{\second}$}
    	\label{fig:results_qual_str1}
    \end{subfigure}%
    \hspace{0.03\textwidth}
    \begin{subfigure}{0.27\textwidth}
        \centering
    	\includegraphics[width=\linewidth, trim = 180 65 0 65, clip]{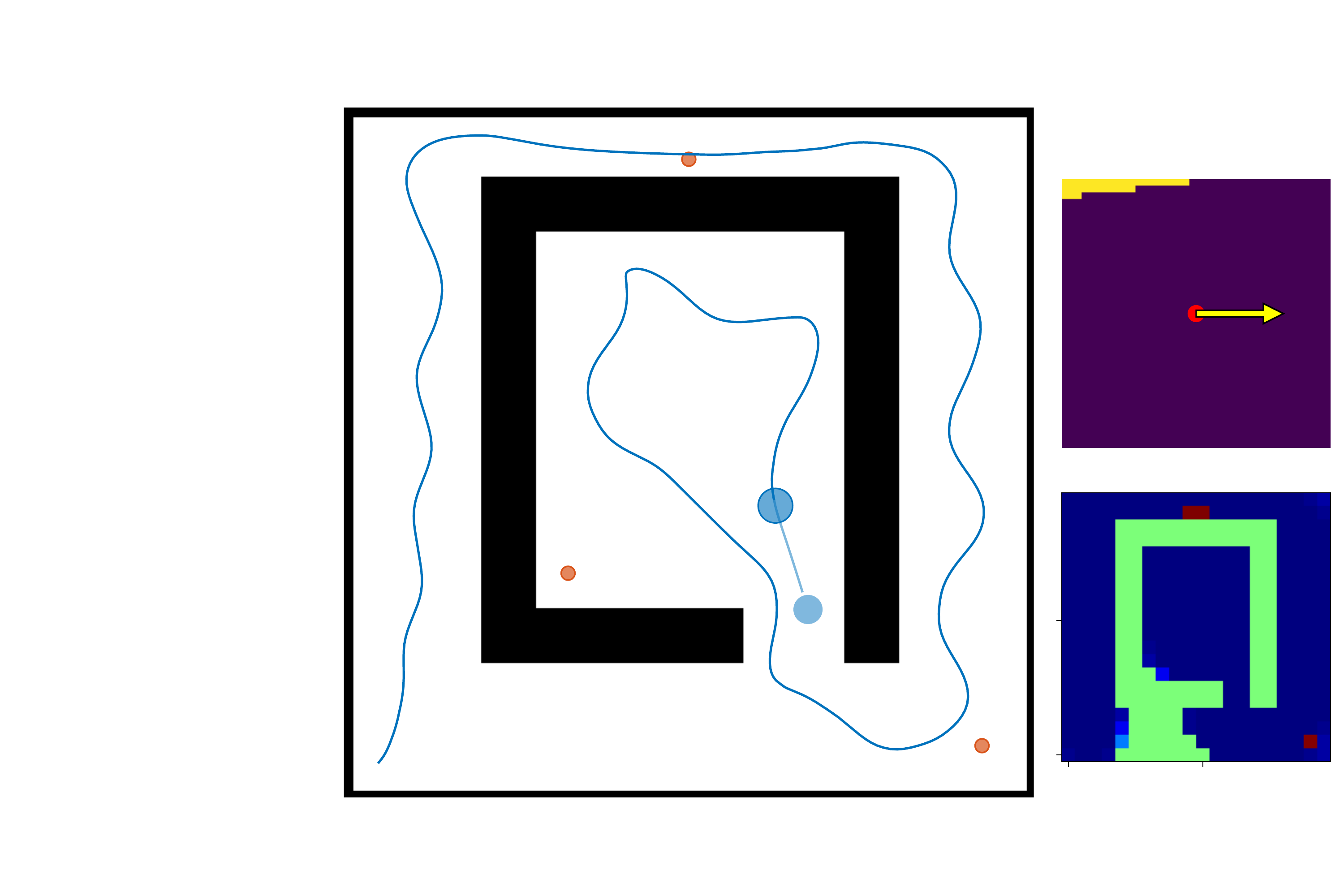}
    	\captionsetup{justification=centering}
    	\caption{$t=\SI{48}{\second}$}
    	\label{fig:results_qual_str2}
    \end{subfigure}%
    \hspace{0.03\textwidth}
    \begin{subfigure}{0.27\textwidth}
        \centering
    	\includegraphics[width=\linewidth, trim = 180 65 0 65, clip]{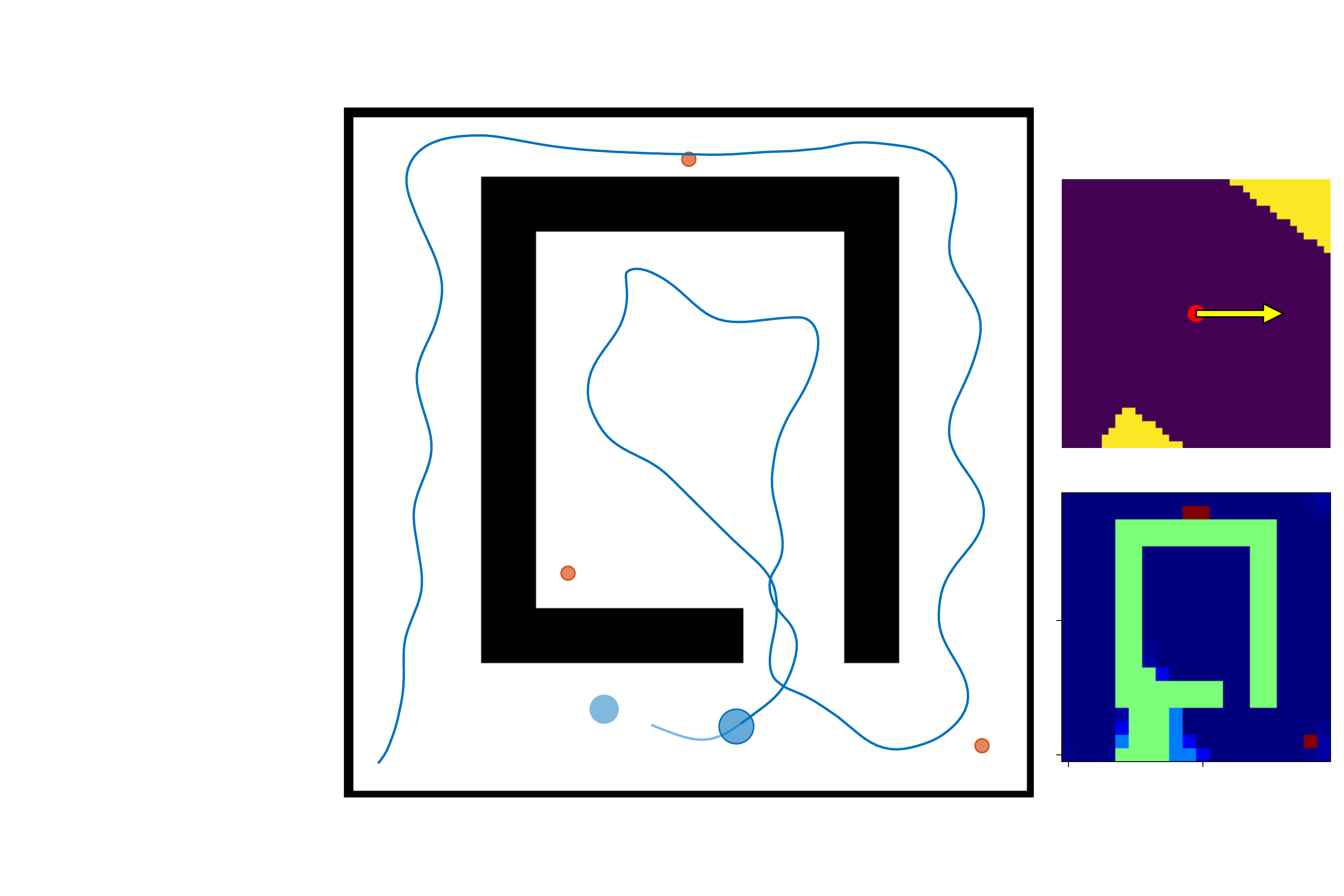}
    	\captionsetup{justification=centering}
    	\caption{$t=\SI{53}{\second}$}
    	\label{fig:results_qual_str3}
    \end{subfigure}%
    \vspace{0.2ex}
    \caption{Trained policy behavior in a \textit{structured} environment of higher complexity than in training, with three timesteps of an episode displayed. The three snapshots show that the robot is effectively guided into and out of a room-like structure, without visiting areas twice. The setup of the figures is as described in \cref{fig:results_qual_unstr}.\looseness=-1}
    \label{fig:qual_structured}
    \vspace{-2ex}
\end{figure*}

We compare the performance against two baseline approaches: one
myopic and one non-myopic informative path 
planning method.  
Similar to our approach, we use both baselines to compute a reference viewpoint $\vp_t^{{\textrm{ref}}}$ for the MPC.


\subsubsection{Myopic Greedy Viewpoint Selection}
\label{sec:4_results_baseline_greedy}
As a myopic baseline, we use a one-step next-best-view planner similar 
to \cite{Gonzalez-Banos2002}.
\pagebreak \noindent
At each time step, we uniformly sample $N_{nbv}=30$
points $\Tilde{\vp}_i, \forall i =1,...,N_{nbv}$ in the policy's action space $\cA$,
and evaluate the objective $I(\mathbf{M};Z(\Tilde{\vp}_i)|z_{0:t})$ for expected
observations $Z(\Tilde{\vp}_i)$ at these viewpoints. The point with 
the highest reward is chosen and passed as $\vp_t^{{\textrm{ref}}}$ to the MPC. This greedy method is also used for warm-starting the training, as explained in \cref{sec:3_method_training}.

\subsubsection{Non-Myopic Monte Carlo Tree Search (MCTS)}
\label{sec:4_results_baseline_mcts}
We use an MCTS planner \cite{Lauri2016, Patten2018, Best2019} as a baseline to find finite-horizon sequences of viewpoints that maximize cumulative information rewards. We build on an open-source Python implementation of Dec-MCTS \cite{Best2019}, and use it for single-robot planning. The planner uses a simplified first-order kinematic model of the robot dynamics and a small set of motion primitives as discrete action space. The motion primitives are combinations from different velocity and angular velocity inputs, $u_{v}$ and $u_\omega$, as given in \cref{tab:hyperparameters}, with a length of $N_a$ timesteps. The planning horizon is $H_{\text{MCTS}}$, for each replanning, $N_{\text{tree}}$ MCTS iterations are performed, and each new leaf node is evaluated with $N_{\text{sim}}$ rollouts. The first position in the best plan is passed as $\vp_t^{{\textrm{ref}}}$ to the MPC. We replan at every time step $t$, but the planning time does not affect the simulated time due to the sequential implementation. Thus the robot does not have to stop for replanning.
Furthermore, our MCTS implementation has access to the global obstacle map for computing rewards of possible future positions. \looseness=-1

\subsection{Performance Results}
\label{sec:4_results_perf}

This section presents the quantitative performance results of our method and the two baselines. The results, summarized in \cref{tab:performance}, are aggregated over a set of random environments for three map complexity levels
defined by the number of sampled obstacles. 
For each number of obstacles, 100 random scenarios are simulated for each of the methods. 
In each episode, the agent has a maximum of $t_\text{max}$ to reach the coverage goal of $\beta=0.9$ before it is considered a \textit{failure}.
We quantify the performance by the average cumulated reward over an episode, the percentage of failure episodes, the average travel time, and the average runtime (excluding the MPC) of the three viewpoint recommending methods. 

Our method outperforms the greedy next-best-view baseline in terms of average episode rewards, completion time, and failures for all map complexities. 
The greedy method exhibits a large number of failures
because it cannot reason about unexplored areas outside the local surroundings.
Thus the robot often revisits already explored areas multiple times instead of moving to unexplored areas to complete coverage. 
Moreover, our method achieves the lowest percentages of failure episodes and the lowest execution times.
Failures of the MCTS planner occur when it determines viewpoint references that are unreachable for the MPC, which our method avoids by training with the MPC. 
The long runtimes of MCTS are caused by the expensive computation of the set of visible cells $\cI_t$ (\cref{sec:2_prelim_belief}) for a large number of viewpoint candidates during planning, necessary to evaluate their information gain. In contrast, our trained policy $\pi$ can infer a promising viewpoint  reference only from currently available observations.
This comes at the cost of suboptimal average rewards and completion time compared to the MCTS planner.
Note, however, the MCTS planner's advantageous assumptions (\cref{sec:4_results_baseline_mcts}), as the long runtime does not affect performance and the global obstacle knowledge enable evaluating rewards for distant positions during planning.

\subsection{Qualitative Analysis}
\label{sec:4_results_qa}

This section analyzes the behavior of our proposed method in two scenarios not used during training and with higher
complexity than the training scenarios, in terms of obstacle placement and an increased coverage goal of $\beta=0.95$. \cref{fig:results_qual_unstr,fig:qual_structured} show the agent path for three different time steps with the recommended viewpoint, the local observation, and belief 
map of the agent for each scenario. In \cref{fig:results_qual_ust1,fig:results_qual_ust2,fig:results_qual_ust3}, the viewpoint reference leads the agent into the most promising unobserved areas and
enables it to enter and leave a narrow  
dead-end corridor at the top of the map. While not globally optimal, the behavior exhibits an efficient strategy of guiding the robot towards unobserved areas, maximizing information gains, and dealing with difficult environment structures. In \cref{fig:results_qual_str1,fig:results_qual_str2,fig:results_qual_str3}, the robot is able to enter and leave a room-like structure. The policy guides the robot to observe inside the room when reaching the entrance, instead of moving further, and decides to leave the room as soon as almost all available information has been gathered. Subsequently, it guides the robot into the remaining unobserved areas. \looseness=-1

\section{Conclusions and Future Work}

In this paper, we introduced a navigation policy capable of guiding a local trajectory planner towards maximizing the information gathered in an unknown environment. We employed reinforcement learning to learn the information-gathering policy using only locally available observations and previously gathered information. The policy learns to maximize information-theoretic rewards by providing a viewpoint reference that an MPC-based local motion planner uses to generate trajectories respecting the robot's safety constraints. The results show that the learned policy is able to effectively guide the robot through unseen environments, and achieve quantitative performance comparable to an MCTS planner. Moreover, our method can be run at a rate three orders of magnitude faster than the MCTS planner, allowing for quick reactions in dynamic scenarios. Future work will consider a limited field of view and experiments on a real robotic platform.  \looseness=-1

\FloatBarrier


\bibliographystyle{IEEEtran}
\bibliography{IEEEabrv, references}


\end{document}